\documentclass[pmlr,twocolumn,10pt]{jmlr} % W&CP article
\usepackage{listings}
\usepackage{float}
\usepackage{lineno}
\usepackage{booktabs}
\usepackage{siunitx}

\theorembodyfont{\upshape}
\theoremheaderfont{\scshape}
\theorempostheader{:}
\theoremsep{\newline}

\jmlrvolume{LEAVE UNSET}
\jmlryear{2026}
\jmlrsubmitted{LEAVE UNSET}
\jmlrpublished{LEAVE UNSET}
\title[Phenotyping based on LLM ]{LLM-Augmented Computational Phenotyping of Long Covid}

 \author{%
 	  \Name{Jing Wang}\Email{jing.wang20@nih.gov}\\
 	\Name{Jie Shen}\Email{jie.shen@stevens.edu}\\
 	 	\Name{Amar Sra}\Email{amarsra@email.gwu.edu}\\
 	\Name{Qiaomin Xie}\Email{qiaomin.xie@wisc.edu}\\
 	\Name{Jeremy C Weiss}\Email{jeremy.weiss@nih.gov}
 }

\begin{document}

\maketitle

\begin{abstract}
Phenotypic characterization is essential for understanding heterogeneity in chronic diseases and for guiding personalized interventions. Long COVID, a complex and persistent condition, yet its clinical subphenotypes remain poorly understood. In this work, we propose an LLM-augmented computational phenotyping framework ``Grace Cycle'' that iteratively integrates hypothesis generation, evidence extraction, and feature refinement to discover clinically meaningful subgroups from longitudinal patient data. The framework identifies three distinct clinical phenotypes, Protected, Responder, and Refractory, based on 13,511 Long Covid participants. These phenotypes exhibit pronounced separation in peak symptom severity, baseline disease burden, and longitudinal dose-response patterns, with strong statistical support across multiple independent dimensions.

This study illustrates how large language models can be integrated into a principled, statistically grounded pipeline for phenotypic screening from complex longitudinal data. Note that the proposed framework is disease-agnostic and offers a general approach for discovering clinically interpretable subphenotypes.% and monitoring disease progression in large-scale observational cohorts.
\end{abstract}

\paragraph*{Data and Code Availability}
This data used for this study is part of the National Institutes of Health's Researching COVID to Enhance Recovery (RECOVER) Initiative, which seeks to understand, treat, and prevent PASC \url{https://recovercovid.org/}, which is available by application. The underlying code for this study is available as supplemental material.
%
%%\paragraph*{Institutional Review Board (IRB)}
%%If your research does not require IRB approval, then you
%%must state this to be the case. 
%
%
\section{Introduction}
\label{sec:intro}
\begin{figure}[t]
	\centering
	\includegraphics[width=0.8\linewidth]{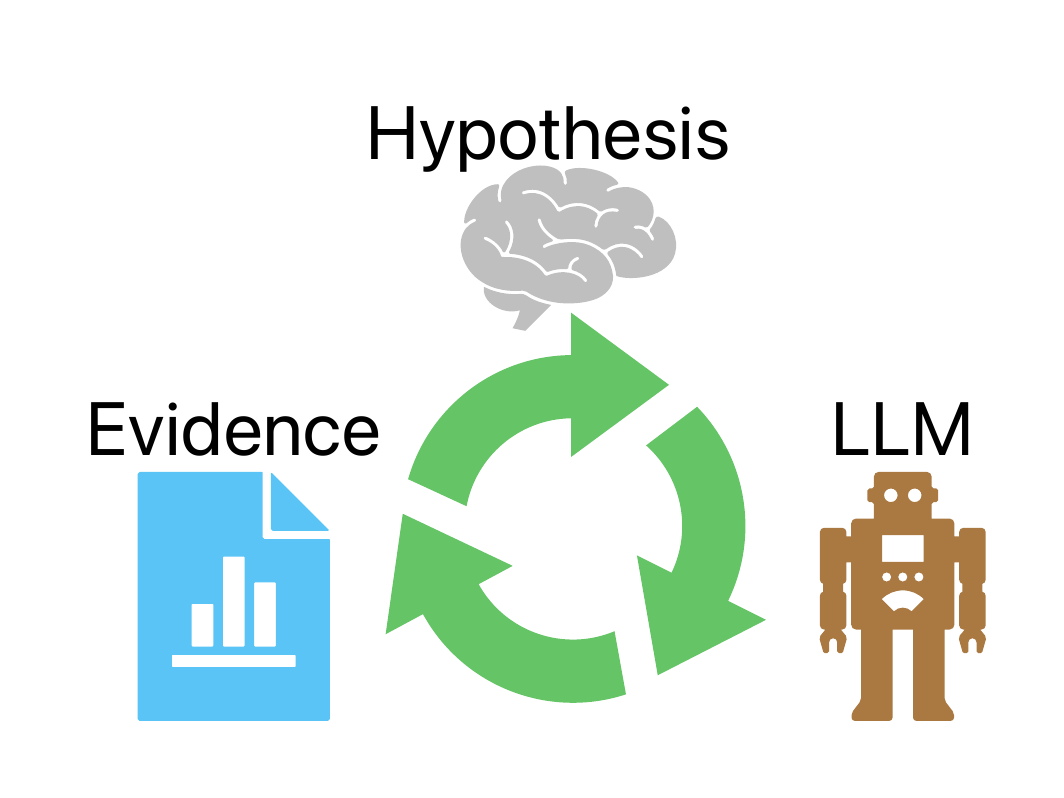}
	\caption{The pipeline of the LLM augmented phyenotyping for Long Covid, we called ``Grace Circle''. We starts with a hypothesis, LLM looks for evidence from the data to support the hypothesis. With feedback about alignment with hypothesis and evidence by LLM, update the data and hypothesis. Repeat the process until the evidence are aligned with hypothesis.}
	\label{fig:sub}
\end{figure}

\begin{table*}[t]
	\centering
	\caption{Group comparisons across Protected, Responder, and Refractory cohorts.}
	\label{tab:group_comparisons}
	\begin{tabular}{lccccc}
		\hline
		Feature & Protected & Responder & Refractory & Test Statistic & p-value \\
		\hline
		Peak PASC Score
		& $4.98 \pm 6.3$ & $18.40 \pm 7.1$ & $22.80 \pm 5.1$
		& $H=4215.2$ & $<0.001$ \\
		
		Initial Score
		& $4.98$ & $8.50$ & $14.20$
		& $F=128.4$ & $<0.001$ \\
		
		Avg.\ Vaccine Doses
		& $3.41$ & $2.38$ & $1.82$
		& $F=84.2$ & $<0.001$ \\
		
		%			Brand (Pfizer/Moderna)
		%			& $31\%/69\%$ & $52\%/48\%$ & $49\%/51\%$
		%			& $\chi^2=214.5$ & $<0.001^{b}$ \\
		\hline
	\end{tabular}
\end{table*}
Phenotying research is widely recognized as fundamentally important in biology, medicine, and genetics. For example, diagnostic phenotyping is primarily used to support the diagnosis of disease \cite{chen2022development}. By analyzing abnormal changes in behaviors of physiological patterns, phenotyping can help identify early signs of a disease, such as sleep pattern changes leads to depression \cite{kas2024digital}.  Predictive phenotyping aims to forecast future health events or risks of diseases, such as monitoring heart rate and physical activity for cardivascular diseases \cite{pavarini2022data}. Monitoring phenotyping is used for ongoing monitoring of diagnosed health conditions or disease progression \cite{langholm2023monitoring}. This work studies belongs to monitoring phenotyping. progression. We study the cohort with Long Covid from RECOVER program.

Digital phenotyping is introduced in 2015 and offers new avenues for research and management of mental heath and physiological disease. It includes collecting health data digitally, supporting early disease diagnosis and health management. The streaming data are from smart devices, sensors, and mobile apps about behavior, psychological, and physiological behavior \cite{huckvale2019toward}. Physiological phenotyping which involves monitoring an individual's physiological paramters to assess their health. The physiological data includes breathing, sleep, heart rate, oxygen level, and activities. We find that breathing is most related to the PASC score. In this work, we use the wearable data of the participants in RECOVER program \cite{zhang2025comprehensive} which includes sleep, respiratory rate, heart rate, and physical activity.

The global COVID-19 pandemic has last more than 4 years. More than 658 million people worldwide have been infected with SARS-CoV-2 (Long COVID) \cite{who}. Long Covid can manifest in people across the life span, from children to older adults. It is a complex disease with sequelae across almost all organ systems. There are many subtypes that may have different risk factors (genetic, environmental, etc.) and distinct biologic mechanisms that may respond differently to treatments \cite{al2024solving}. It is a major and ongoing public health challenge, with substantial impacts on quality of life and health-care utilization \cite{danesh2023symptom,huang2021retracted,grady2025impact,wang2025active}.  The observational analyses are very important. For example, it is suggested that use of the antiviral ritonavir-boosted nirmatrelvir within 5 days of symptom onset of SARS-CoV-2 infection may reduce the risk of Long Covid by 26\% \cite{xie2023association}. The diabetes drug metformin initiated within 7 days of SARSCoV-2 infection reduced the risk of Long Covid by 41\% in a RCT \cite{bramante2023outpatient}. More evidence is needed to evaluate the effectiveness of reducing Long Covid risk and the safety of various antivirals. It is also important to understand different subtypes of Long Covid.

Most recently, the deep learning tools has and AI tools have been used for healthcare related research, such as phenotyping. genetics, clinical trails \cite{schmidt2026automating}. In this work, we leverage large language models (LLMs)  to analyze detailed clinical profiles of Long COVID participants enrolled in the NIH RECOVER Program.

\begin{figure*}[h!]
	\centering
	\includegraphics[width=0.95\linewidth]{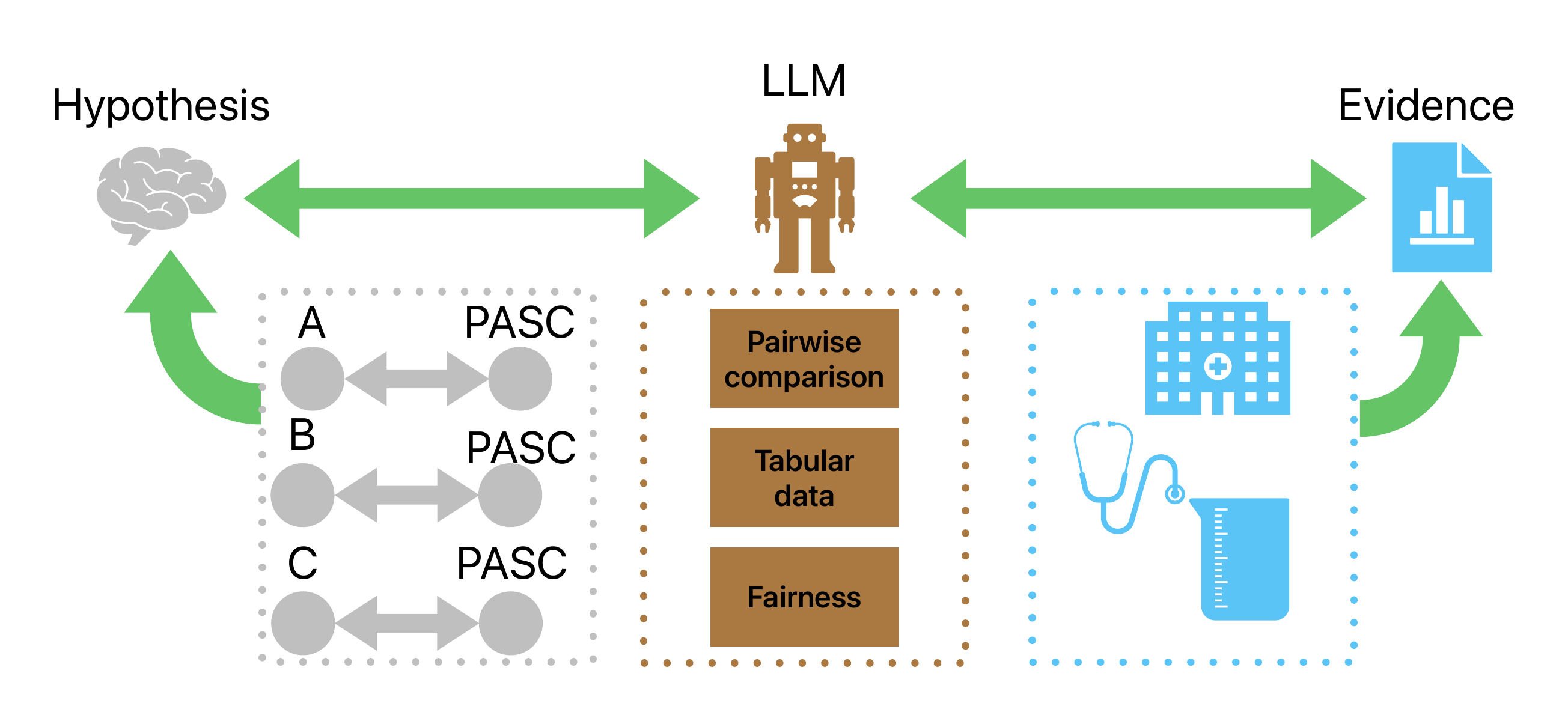}
	\caption{The steps in the pipeline. LLM performs pairwise comparison, fairness criteria in the tabular data reading. The hypothesis updates by choosing different hypothesis from the pool. Evidence is updated by collecting related features, replacing or removing unrelated features from the knowledge base. The process continues until hypothesis and evidence are aligned.} 
	\label{fig:end}
\end{figure*}

\paragraph{Main result.} We propose the pipeline to use LLM for auto clinical phenotyping as shown in Figure \ref{fig:end}.  Given the participants' data with Long Covid, we first propose the weak assumption, then LLM reads the time series data of participants to seek evidence to support the hypothesis. If evidence does not support hypothesis, we collect more features of the participants, revise the hypothesis, and let LLM find evidence again. The process converges when the evidences support the hypothesis. Then statistical analysis is conducted to verify the significance of alignment. It is related to the most recent works about LLM agent that can be updated automatically. However, due to the complexity of the task in clinical domain, it requires human involvement for revising the hypothesis. It is a general pipeline. If we replace the Long Covid data to another disease, then the final hypothesis would be new idea or discoveries that applies in that domain. For example, if the data becomes ICU patient discharge summaries, the hypothesis converges to the discoveries of ICU patient. 

\paragraph{Findings.} In the analysis of data from 13,511 participants in the RECOVER adult cohort, a prospective longitudinal cohort study. With PASC score as outcome, which is the evaluation of 44 symptoms (postexertional malaise, fatigue, brain fog, dizziness, gastrointestinal symptoms, palpitations, changes in sexual desire or capacity, loss of or change in smell or taste, thirst, chronic cough, chest pain, and abonormal movements), we set PASC less than 12 as State 0 and above 12 as State 1, as 12 is the suggested severity threshold \cite{thaweethai2023development}. We identify three clinical subpheynotyping based on the response of the vaccines. 
\begin{itemize}
	\item Protected. 9,544 individuals (69\% female; 8\% Asia, 16\% Hispanic/Latino; 17\% non-Hispanic Black; median age, 44 years [IQR, 33-60]) maintaining Status 0 throughout the study.
	\item Responder: 3,302 individuals (76\%  female; 6\% Asia; 18\% Hispanic/Latino; 14\% non-Hispanic Black; median age, 49 years [IQR, 37-60]) who exhibited a state transition from Status 1 to Status 0.
	\item Refractory. 665 individuals (82\% female; 5\% Asia; 16\% Hispanic/Latino; 8\% non-Hispanic Black; median age, 48 years [IQR, 39-57]) maintaining Status 1 regardless of intervention.
\end{itemize}
Then we conduct the statistical analysis across multiple independent dimension, symptom severity, baseline status, treatment intensity to provide strong convergent validity for the discovered subphenotyping as shown in Table 	\ref{tab:group_comparisons}. 
\begin{itemize}
	\item Peak PASC severity differs dramatically across subphenotypes. The Protected group shows a low mean peak PASC score, while the Responder and especially the Refractory groups demonstrate substantially higher symptom burden. The  extremely large Kruskal-Wallis statistic ($H=4215.2$, $p<0.001$) indicates that these differences not only statistically significant but also reflect a pronounced separation in symptom trajectories.
	\item Baseline severity prior to vaccination increases monotonically from Protected to Responder to Refractory. the large F-statistic ($F=128.4$, $p<0.001$) supports the interpretation that the subphenotypes reflect differential vaccine-associated protection, rather than a uniform response across participants.
\end{itemize}
The results indicate that the subphenotyping captures clinically meaningful and biologically relevant structure in Long Covid, with direct implications for risk stratification, prognosis, and personalized intervention strategies.
\section{Related works}

Phenotype research can be classified as the following categories based on applications scenarios, such as public health phenotyping and personalized health phenotyping \cite{zhang2025comprehensive}. Our work belongs to public health phenotyping by aggregating and analyzing the long term followup of a large number of individuals, 13,511. Based on analysis objectives, phebotyping research can be categorized as diagnostic phenotyping, predictive phenotyping, preventive phenotyping and monitoring phenotyping. Our research belongs to the monitoring phenotyping. We continuously monitor the participants with Long Covid their vaccine level, sleep, heart rate, lab test, aiding in Long Covid disease management. Based on the data source, phenotyping research can be mapped  to behavioral phenotyping, physiological phenotyping, psychological phenotyping, environmental phenotyping, social phenotyping, and medical phenotyping. This work mainly collects behavioral data, medical data, physiological data, and psychological data. The outcome in our work is PASC score which is based on 44 symptoms, such as brain fog, dizziness, chronic cough to name a few \cite{thaweethai2023development}.

Deep learning models, Large Language Models are powerful tools on phenotyping research and genomic measurements  \cite{avsec2026advancing, schmidt2024next,schmidt2026automating,chen2025benchmarking}. For example, GPT-4 has been used for sub-phenotyping of patients with Crohn's disease, considering age at diagnosis and disease behavior \cite{schmidt2026automating}.  LLMs may offer an alternative to traditional bioinformatics methods to prioritize disease-associated genes based on disease phenotypes. Therefore, LLM based methods potentially enhance diagnostic accuracy and simplify the process for rare genetic diseases. GPT 3.5 has been used  \cite{peng2024large}. To protect our data are human related data, we use a local Large Language Model for our research.

To solve complex tasks, prompt engineering has been demonstrated as an effective strategy , such as chain-of-thought prompting \cite{wei2022chain,brown2020language}. However, chain-of-thought prompting requires a series of intermediate reasoning steps. In our case to explore the subphenotyping of Long Covid, there is not predefined steps. There are many works about reinforcement learning with human feedback (RLHF) \cite{ouyang2022training,bai2022training,zheng2023judging}. In this work, we propose an auto feedback by comparing the evidence from LLM and hypothesis. Then we require update the hypothesis and data processing to feed LLM with more related data.  Different from existing works that provide all information at once, we provide multiple iterations of data processing, hypothesis update and feedback collection. 

Pairwise comparison. Applications such as the disease risk, treatment recommendation, it is important to consider the relationship between patients, like two patients with similar symptoms tend to recover with similar treatment \cite{zhugradient}. In machine learning, pairwise comparisons have been used in preference learning in recommender systems, ranking and crowdsourced learning \cite{xu2020simultaneous,gong2022ranksim,zeng2022efficient}. This work focuses on learning threshold functions with pariwise comparisons \cite{hopkins2020power,zeng2022list}. By adding weak distributional assumptions and allowing comparison queries, it makes the learning algorithm requires exponentially fewer samples. In this work, we follow the line propose weak assumption first, then feed the similar trajectories of Long Covid to LLM to verify the hypothesis.

\begin{figure}[h!]
	\centering
	\includegraphics[width=0.95\linewidth]{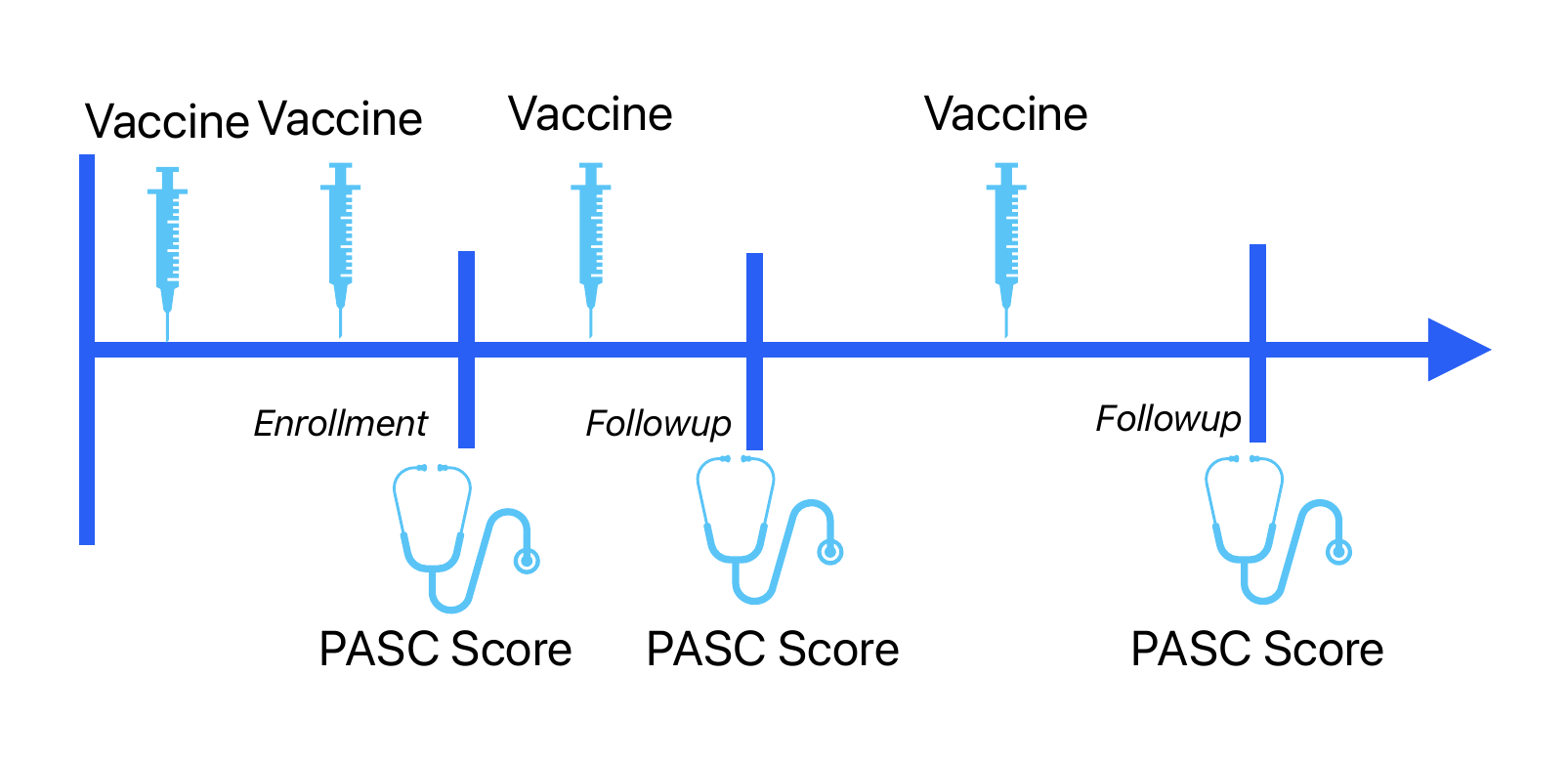}
	\caption{The vaccine and wellness (PASC score) trajectory for Long Covid participants.}
	\label{fig:data}
\end{figure}

\section{Method}

Formally, our data consists of several groups of features $\{F_i\}_{i=1}^D$, where $F_i = \{f_1,\cdots, f_{D_i}\}$ is the $i$th group of features with $D_i$ entities. For example, Figure \ref{fig:data} shows the groups of vaccine record features and the PASC score outcome features. For vaccine record feature, each feature $f_i$ consists of two items $\{e_i, t_i\}$ where $e_i$ is the index of the vaccine, and $e_i$ is the recorded time of the event. For example, $e_i = $ 1st dose, $t_i=$2022-02-20. For the target feature PASC score, each feature $f_i$ consists of two items $\{e_i, t_i\}$ where $e_i$ is the PASC score, and $e_i$ is the recorded datetime of the event. We also have wearable features, For example, $e_i = $ weekly breathing rate, $t_i=$2022-05-20. There are other features, such as weekly Breathing Rate and related summary date. Besides time series features, we also have statistic demographics features, such as sex and race.

The goal of our project is to phenotyping of Long Covid to monitor the progression of Long Covid. The recovery of Long Covid is evaluated by the PASC score. To this end, we first need to find the features that are most related to the PASC score. Then based on that, we find the subphenotyping based on most related features. The end-to-end-pipeline is shown in Figure \ref{fig:end}. We will introduce each component in the pipeline in the following section.

\paragraph{Hypothesis}Then we initialize a list assumptions $H$. For example, $h_0 =$ does the Long Covid recover over time?, $h_1=$ does the Long Covid participants recover after Covid booster?, $h_2=$Is Long Covid symptoms related to breathing? The hypothesis is updated based on the LLM analysis of the features.

\paragraph{Pairwise comparison} Given the selected feature subset $S=\{F_1, \cdots, F_d\}$. We compute the similarity between the participants in two ways. First, we compute the similarity based on the statistical analysis, such as the number of recorded vaccines / PASC Sore taken so far. Then we compute the semantic similarity by LLM based the embedding, such as MedGemma, Qwen3. Given the similarity between the participants, we feed the samples with certain feature subset $S$ individually, and with the top $k$ similar pairs and in batches. 

\paragraph{Fairness} Machine learning models have shown biased predictions again disadvantaged groups on several real-world tasks \cite{dressel2018accuracy,chaialignment,shen2022metric}. Similar to accuracy, fairness can also be targeted by malicious adversarials, leading to biased outcomes against certain demographics. There are some techniques, including preprocessing that is to adjust training distribution to reduce discrimination \cite{jiang2020identifying}; in-processing to impose fairness constraint during training by reweighing or adding relaxed fairness regularization \cite{jung2025adversarial}; and post-processing to adjust the decision threshold in each sensitive group to achieve expected fairness parity \cite{cruz2023unprocessing}. In this work, we use proprocessing to guarantee the fairness, first we filter demongraphic features in fairness comparison. The similarity threshold between participants are computed with protected features. We also sample the data randomly multiple times to confirm the alignment between evidence and hypothesis.
\begin{figure*}[h!]
	\centering
	\includegraphics[width=0.95\linewidth]{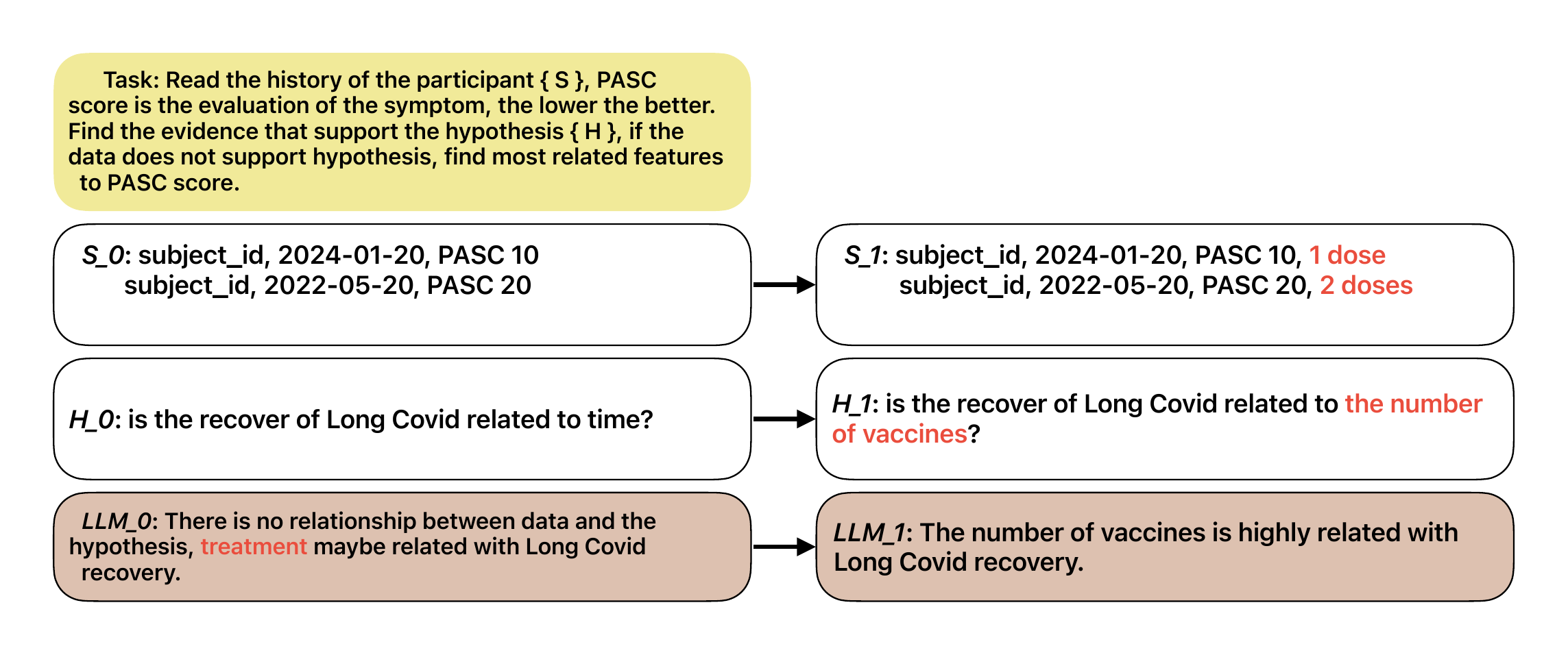}
	\caption{The update of hypothesis and feature space given the Long Covid cohort. The target variable is the PASC score which reflects the overall wellness of the participants of Long Covid. } 
	\label{fig:prompt}
\end{figure*}
\paragraph{Group Feature Selection}  If LLM finds evidence in the given feature subset to support the hypothesis, then the process terminates, we conduct statistical analysis to verify the results. If the provided feature subset does not support the hypothesis, we have two actions. First is to revise the hypothesis based on the insights from LLM. The second is to update the feature subset by selecting hypothesis related features and removing unrelated features. For example, the LLM discovers that weekly breathing rate is more related to the PASC score, then in the next round, we update the hypothesis to be ``how weekly breathing rate is related to PASC score over time?'' The feature subset is updated by removing unrelated features, such as heart rate, activity, and include breathing related features, such as REM sleep breathing rate, all time/weekly/monthly breathing rate.

 \paragraph{LLM judgment} Given a hypothesis $h$, we request certain a subset from $\{F_i\}_{i=1}^D$ and find evidence to support the hypothesis. We use the thinking mode of the Large Language Models. If the features are aligned with the hypothesis, then LLM outputs the evidence from the features. If the features are not aligned with the hypothesis, LLM provides feature candidates that may be related to hypothesis. Figure \ref{fig:prompt} provides an example of the hypothesis and feature update process. The first hypothesis is assuming recover is related over time. But based on the LLM analysis result, there is no direct correlation. LLM suggests to include treatment, then we choose vaccine as treatment. The updated hypothesis becomes the relationship between number of vaccines so far and recovery. Then the evidence supports the hypothesis. The iteration terminates. In practice, we could also include more features at first, and let LLM remove unrelated features. The example in Figure \ref{fig:prompt} includes low dimensional features first, which allows LLM to read as many participants' records as possible. Given the limit of the maximal lengths of input of LLM, there is a trade off between  the number of records and the dimension of features. 

\section{Experiments}

\section{Dataset}

Our dataset is a subgroup of the RECOVER adult cohort with adult participants  enrolled before April 10, 2023. The analysis cohort included participants with a study visit completed 6 months or more after the index date \cite{horwitz2023researching}. The participants are from 86 sites in 33 U.S. states, Washington, DC and puerto Rico, via facility and community-based outreach. The participants complete quarterly questionnaires about symptoms, social determinants, vaccination stauts, and interim SARS-CoV-2 infections. In addition, participants contributes biospecimens and undergo physical and laboratory examinations at approximately 0, 90 and 180 days from infection or negative test date, and yearly thereafter. The primary outcome is onset of PASC, measured by signs and symptoms. 

The experiments are conducted on an AWS p3 instance with 8 GPUs. 

\subsection{LLM Judgment}

Given our longitudinal data, we allow the LLM to read and summarize the evidence directly. We use pairwise comparisons to jointly process multiple participants at a time (e.g., 4, 10, or 20 participants per batch). Through this process, the LLM identifies distinct patterns: some participants consistently exhibit low PASC scores, some consistently high scores, and others show substantial fluctuations over time.

\begin{figure*}[h!]
	\centering
	\includegraphics[width=0.95\linewidth]{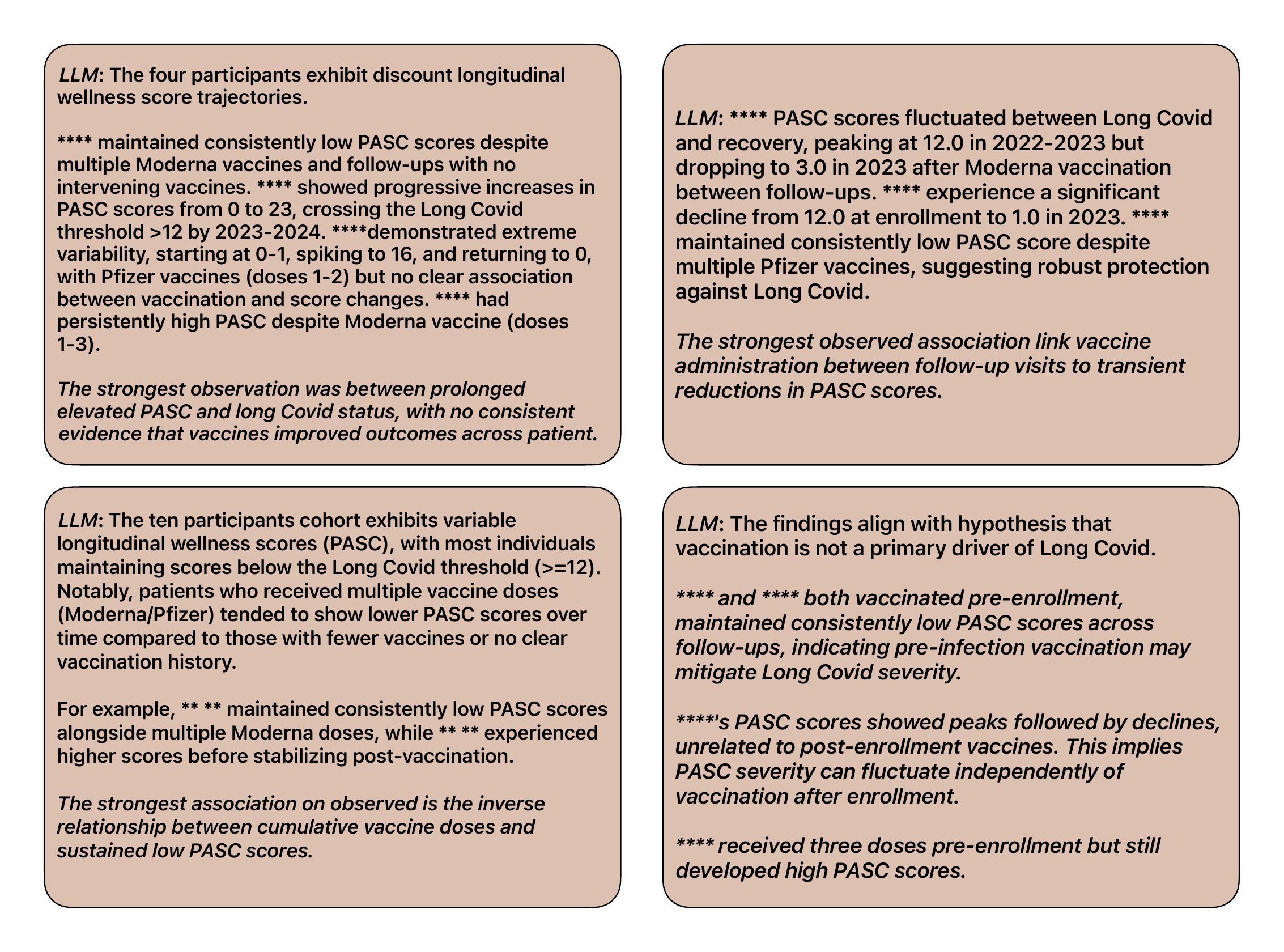}
	\caption{Example of insights of LLM after reading  trajectories of multiple participants. } 
	\label{fig:result}
\end{figure*}

Based on these observations, we examine the correlation between vaccination and recovery. As shown in Figure \ref{fig:result}, individual vaccine events are not strongly associated with PASC scores. We then refine the analysis by introducing an additional feature, the cumulative number of vaccine doses. The LLM identifies dose 5 as a potential threshold beyond which a significant difference in PASC scores emerges.

Finally, using the pipeline outputs on the selected samples, we construct a dataset consisting of time-series PASC scores and vaccination records, and conduct downstream statistical analyses.

%\section{Statistical Analysis}
%
%\paragraph{Survival analysis}
%
%
%\paragraph{Pears correlation} 

\begin{table*}[t]
	\centering
	\caption{\textbf{Demographic characteristics.}}
	%	\label{tab:demographic_characteristics}
	\begin{tabular}{lccc}
		\hline
		\textbf{Characteristic} & \textbf{Protected} & \textbf{Responder} & \textbf{Refractory} \\ \hline
		{\#  Participants} &9,544 & 3302 & 665 \\ \hline
		%		{\#  Observations} &62,060 &28,085 & 4,086 \\ 
		%		\hline
		Age, median (IQR) & 44.0 (33.0-60) & 49.0 (37-60) & 48.0 (39-57) \\
		\hline
		Age category at enrollment &  &  &  \\
		\quad 18-45 & 32\% & 41\% & 48\% \\
		\quad 46-64  & 52\% & 44\% & 43\% \\
		\quad $>$65 & 16\% & 15\% & 9\% \\
		\hline
		Sex assigned at birth &  &  &  \\
		\quad Female & 70\% & 76\% & 82\% \\
		\quad Male & 30\% & 24\% & 18\% \\ \hline
		%		\quad Intersex & 4 & 1 & 0 \\
		%		\quad Unknown & 33 & 3 & 1 \\
		Race & & & \\
		\quad Asian & 8\% & 6\% & 5\%\\
		\quad Hispanic and Latino & 16\% & 18\% & 16\% \\
		\quad non-Hispani Black & 17\% & 14\% & 8\%  \\
		\hline
	\end{tabular}
	\label{tab:data}
\end{table*}
\section{Findings with Statistical Analysis}

\subsection{Baselines}
we compared our approach against two standard methods for longitudinal data analysis: Linear Mixed-Effects Models (LMM) and Latent Class Trajectory Modeling (LCTM).

We initially fit an LMM to evaluate general temporal trends and quantify population-level variance. The model yielded a fixed time effect of $\beta=0.010$ and $p$-value=0.017 and a random group variance of $\sigma^2=30.01$. The magnitude of the group variance quantitatively confirms substantial between-subject heterogeneity within the dataset. This high degree of variance indicates that modeling the cohort via a single, population-averaged longitudinal trajectory is insufficient, necessitating the extraction of distinct patient subgroups.

To derive data-driven patient subgroups, we applied LCTM using the StepMix algorithm \cite{morin2025stepmix}, specifying a 3-class model. The resulting mixture model partitioned the patient trajectories as follows:
\begin{itemize}
	\item Class 1 (29.7\%): Characterized by a low initial baseline followed by a slightly decreasing trajectory over time.
	\item Class 2 (37.9\%): Exhibited a persistently high and stable trajectory throughout the observation window.
	\item Class 3 (32.4\%): Demonstrated a low initial baseline with a distinctly sharper subsequent decline compared to Class 1.
	\end{itemize}
While the morphological shapes of these latent classes broadly align with the phenotypes identified by our approach, the traditional LCTM baseline exhibited significant optimization limitations. Most notably, the StepMix model failed to achieve full convergence, and the covariance matrix for one of the extracted classes collapsed, exhibiting near-zero variance. This parametric instability suggests that traditional trajectory clustering is ill-suited for the structural complexities of this clinical setting. In contrast, our framework circumvents these convergence failures and class-collapse issues, yielding phenotypes that are both mathematically stable and highly interpretable.

\subsection{Subphenotyping}

\begin{figure*}[h!]
	\centering
	\begin{minipage}[t]{0.49\linewidth}
		\centering
		\includegraphics[width=\linewidth]{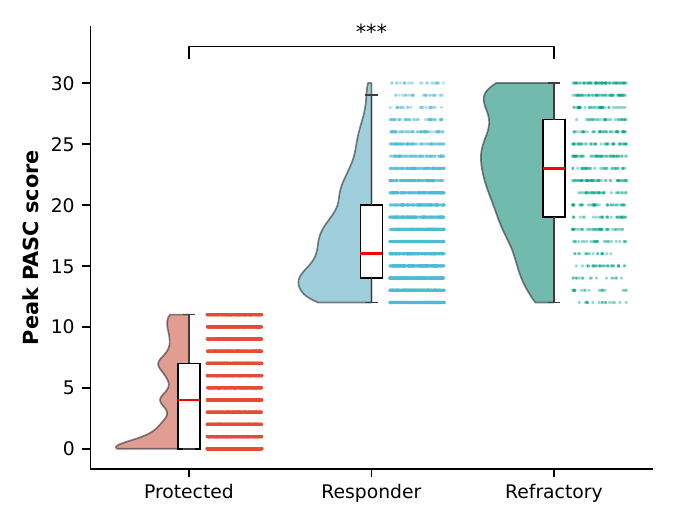}
		\textbf{A}\par
	\end{minipage}\hfill
	\begin{minipage}[t]{0.49\linewidth}
		\centering
		\includegraphics[width=\linewidth]{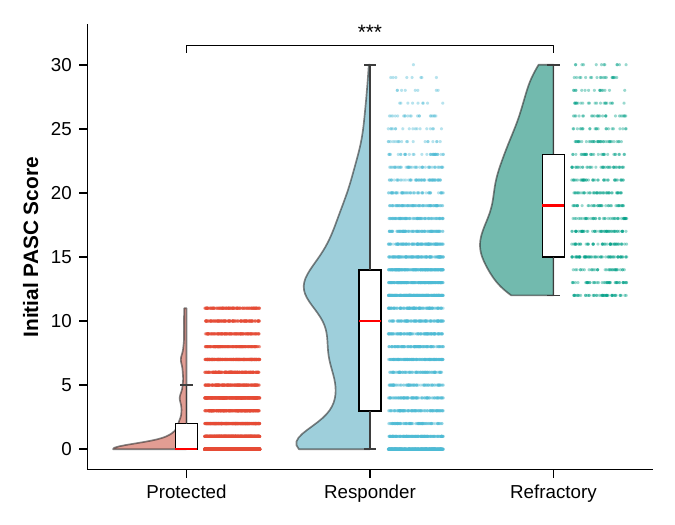}
		\textbf{B}\par
	\end{minipage}
	\caption{\textbf{Distribution of PASC severity across patient cohorts.} \textbf{A,} Distribution of peak PASC severity and individual score variance. \textbf{B,} Distribution of initial PASC severity and individual score variance.}
	\label{fig:peak_initial}
\end{figure*}

We discover 3 distinginshed  Subphenotyping. The protected cohort with consistent low PASC score. The Responder with at least one status transfer from 1 to 0. The Refractory with consistent high PASC score. The demographic characteristics of three are shown in Table \ref{tab:data}.  Figure \ref{fig:peak_initial} (A) and (B) plots illustrates the distribution of peak and initial PASC scores across the three identified clinical cohorts: Protected, Responder, and Refractory. 

\paragraph{The Kruskal--Wallis test} is performed at the participant level using peak PASC severity. 
For each participant, we extract the peak PASC score observed across all recorded time points. 
Each participant has a longitudinal record consisting of the observation date, the corresponding PASC score, 
and the cumulative number of vaccine doses received by that date. 
Participants are then assigned to one of three groups.

Let $n_i$ denote the number of participants in group $i$, for $i \in \{1,2,3\}$, and 
let $R_i$ denote the sum of ranks of peak PASC scores in group $i$ after pooling all participants. 
The Kruskal--Wallis $H$ statistic is computed as
\[
H = \frac{12}{N(N+1)} \sum_{i=1}^{3} \frac{R_i^2}{n_i} - 3(N+1),
\]
where $N = \sum_{i=1}^{3} n_i$ is the total number of participants. 
Under the null hypothesis that the distributions of peak PASC scores are identical across the three groups, 
$H$ asymptotically follows a $\chi^2$ distribution with $2$ degrees of freedom. The  effect size is 0.63.
In our analysis, the test yields $H = 4215.2$ with a corresponding $p$-value $< 0.001$. It confirms that these cohorts represent distinct clinical trajectories of the disease.

% Protected Cohort demonstrates the lowest peak severity, with a mean score of $4.98 \pm 6.3$. Responder cohort exhibits intermediate severity ($18.40 \pm 7.1$) but shows significant recovery potential post-vaccination. Refractory cohort displays the highest peak PASC scores ($22.80 \pm 5.1$) and persistent symptoms. The significance bracket indicates a highly significant difference between the groups ($p < 0.001$, Kruskal-Wallis $H$-test), confirming that these cohorts represent distinct clinical trajectories of the disease.

\paragraph{Stability of subphenotypes} Using 100 bootstrap samples and Jaccard similarity, we get
\begin{itemize}
	\item Subphenotype Protected: 0.970
	\item Subphenotype Responder: 0.994  
	\item Subphenotype Refractory: 0.972
\end{itemize}
All exceed the 0.85 threshold for strong stability \cite{hennig2007cluster}.

\subsection{Dose Response}

Our analysis reveals that recovery is not a spontaneous event but is closely tied to cumulative vaccine exposure, particularly within the ``Responder'' phenotype.

\begin{figure}[h!]
	\centering
	\includegraphics[width=0.8\linewidth]{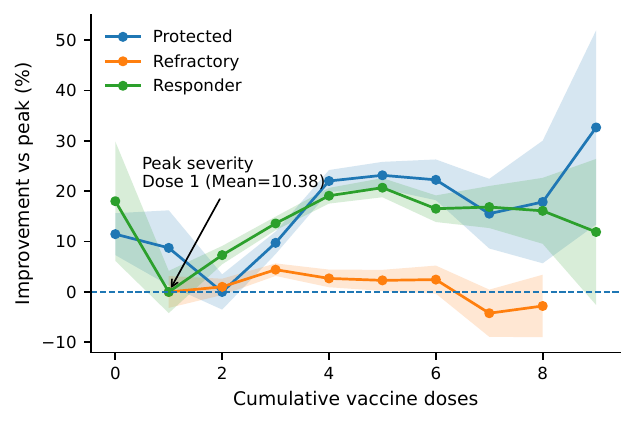}
	\caption{Longitudinal dose-response of PASC severity differs by clinical phenotype.}  %Mean PASC score ($\pm$95\% CI) is plotted against cumulative prior vaccine doses for Protected, Responder, and Refractory cohorts. The Responder group exhibits a clear post-immunization peak at Dose 1 followed by a predictable recovery, reaching $\sim$20.7\% improvement by Dose 5 relative to the peak. The Protected group maintains consistently low symptom burden with modest additional improvement at higher doses. In contrast, the Refractory group shows little early benefit and a late increase in symptom severity at higher cumulative doses, indicating persistent and worsening clinical burden.}
\label{fig:reponse}
\end{figure}
%\subsection{Responders exhibit a reproducible post-immunization peak followed by dose-dependent improvement}
Following an initial post-immunization peak at Dose~1 (mean PASC $=10.38$), Responders followed a consistent recovery trajectory with increasing cumulative doses (Figure~\ref{fig:reponse}). 
From Dose~2 through Dose~5, each additional dose corresponded to an average reduction of $\sim0.54$ PASC points, reaching a 20.71\% improvement by Dose~5 relative to the peak ($P<0.001$; Dose~2: $\approx+7.31\%$; Dose~5: $\approx+20.71\%$). 
The 95\% confidence band indicates stable estimation at common dose strata and widening uncertainty at sparsely populated extremes.

In contrast, the protected cohort exhibited low baseline symptom burden with a shallow dose-response. PASC severity peaked at Dose~2 (mean $=1.79$) and declined through Dose~5 (mean $=1.37$), corresponding to a $23.17\%$ improvement relative to the post-immunization peak (average reduction $\approx 0.14$ points per additional dose from 2 to 5; $P<0.001$). At higher cumulative doses, mean PASC continued to trend downward (Dose~9: mean $=1.20$, $\approx+32.67\%$; Dose~10: mean $=1.08$, $\approx+39.70\%$), although estimates beyond Dose~8 were based on smaller sample sizes.

The refractory cohort showed minimal early improvement and a subsequent worsening pattern. After a modest peak at Dose~1 (mean $=20.54$), PASC scores decreased to a shallow nadir at Dose~3 (mean $=19.63$; $\approx+4.43\%$ improvement; $P=0.011$) but remained near $\sim$20 through Dose~6 ($\approx+2.46\%$). At higher cumulative doses, symptom severity increased markedly (Dose~9: mean $=24.20$, $\approx-17.84\%$; Dose~10: mean $=24.43$, $\approx-18.95\%$ relative to the Dose~1 peak), consistent with persistent and worsening clinical burden.

%Although effect sizes were small, 
The consistency across the full cohort and phenotype-defined subgroups supports an interpretation in which elapsed time alone does not correspond to spontaneous symptom resolution, while cumulative immunological exposure is associated with reduced symptom burden in dose-sensitive phenotypes.

\subsection{Time vs dose}

\begin{figure}[h!]
	\centering
	\includegraphics[width=\linewidth]{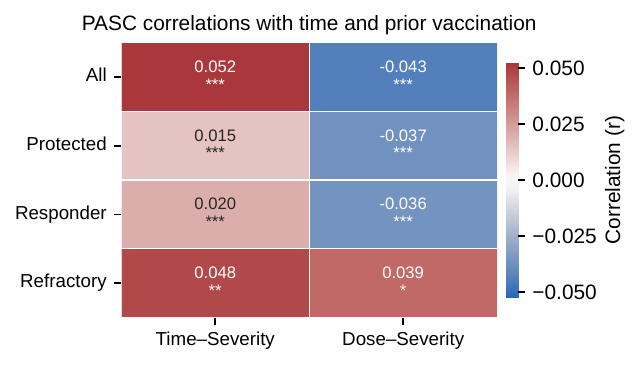}
	\caption{PASC severity shows weak but significant correlations with time and prior vaccination across phenotypes.}
	\label{fig:heatmap}
\end{figure}

To separate passive temporal trends from immunological exposure, we modeled PASC severity against (i) elapsed time since vaccination and (ii) cumulative vaccine dose count. 
We have the conclusion that across all observations, severity increased modestly with time ($r=0.0521$, $P=1.26\times10^{-59}$), whereas cumulative vaccination showed an inverse association with severity ($r=-0.0434$, $P=5.95\times10^{-42}$) (Figure~\ref{fig:heatmap}).

\paragraph{Pearson correlation coefficients} ($r$) between PASC severity and (i) time (Time-Severity) and (ii) prior vaccine dose count across the full cohort and phenotype-defined subgroups are summarized in Figure \ref{fig:heatmap}. Cell color encodes the direction and magnitude of $r$ (diverging scale centered at 0), and each cell is annotated with the corresponding r value and statistical significance. Time-Severity correlations are positive in all groups (All: $r$=0.052; Protected: 0.015; Responder: 0.020; Refractory: 0.048), indicating slightly higher severity with time. Dose-Severity correlations are negative in All, Protected, and Responder (All: r=-0.043; Protected: -0.037; Responder: -0.036), consistent with a modest reduction in severity with higher prior vaccine dose count, while the Refractory cohort shows a small positive Dose-Severity association (r=0.039). Significance is denoted as $p<0.05 (*)$, $p<0.01 (**)$, and $p<0.001 (***)$.

\section{Conclusion}
We present a general, iterative pipeline ``Grace Cycle'' that leverages large language models (LLMs) for automated clinical phenotyping from longitudinal patient data. 
By treating hypothesis generation, evidence extraction, and feature refinement as an interactive loop, the proposed framework enables LLMs to surface latent structure in complex clinical trajectories while remaining grounded in statistical validation. 
Applied to the RECOVER adult cohort, this approach identifies three clinically interpretable PASC subphenotypes—Protected, Responder, and Refractory, characterized by distinct symptom trajectories and differential responses to vaccination.
The strong separation across peak symptom severity, baseline status, and treatment intensity provides convergent evidence that these subphenotypes capture meaningful heterogeneity in Long COVID progression.
More broadly, the framework is disease-agnostic and can be transferred to other clinical domains, where it may facilitate hypothesis discovery and accelerate data-driven clinical insights.

\section{Limitations.}
This study has several limitations.
First, although the LLM-guided pipeline automates evidence discovery and feature exploration, human expertise remains essential for hypothesis revision and clinical interpretation, particularly in high-stakes medical settings.
Second, the analysis is observational in nature; therefore, the identified associations between vaccination and PASC trajectories should not be interpreted as causal.
Residual confounding factors, such as healthcare access, comorbidities, or unmeasured behavioral variables, may influence observed outcomes.
Third, PASC severity is derived from self-reported symptoms, which are subject to recall bias and reporting variability.
Finally, while the proposed framework is general, its performance and interpretability may depend on the quality and granularity of longitudinal data available in other disease domains.
Future work will focus on incorporating causal modeling, improving robustness to noisy clinical records, and reducing the need for human intervention in hypothesis refinement.
%\section{Citations and Bibliography}
%\label{sec:cite}
%
%The \textsf{jmlr} class automatically loads \textsf{natbib}
%and automatically sets the bibliography style, so you don't need to
%use \verb|\bibliographystyle|.
%This sample file has the citations defined in the accompanying
%BibTeX file \texttt{chil-sample.bib}. For a parenthetical
%citation use \verb|\citep|. For example
%\citep{guyon-elisseeff-03}. For a textual citation use
%\verb|\citet|. For example \citet{guyon2007causalreport}. 
%Both commands may take a comma-separated list, for example
%\citet{guyon-elisseeff-03,guyon2007causalreport}.
%
%These commands have optional arguments and have a starred
%version. See the \textsf{natbib} documentation for further
%details.\footnote{Either \texttt{texdoc natbib} or
%\url{http://www.ctan.org/pkg/natbib}}
%
%The bibliography is displayed using \verb|\bibliography|.
%
%\acks{Acknowledgments go here \emph{but should only appear in the
%camera-ready version of the paper if it is accepted}.
%Acknowledgments do not count toward the paper page limit.}
%
\bibliography{chil-sample}
\bibliographystyle{natbib}
%
%\appendix
%
%\section{First Appendix}\label{apd:first}
%
%This is the firs\bt appendix.
%
%\section{Second Appendix}\label{apd:second}
%
%This is the second appendix.

\end{document}